\documentclass[10pt]{IEEEtran}

\ifCLASSINFOpdf

\fi

\usepackage[subrefformat=parens,labelformat=parens,caption=false,font=footnotesize]{subfig}
\usepackage{array}
\usepackage{balance}
\usepackage{amsmath}
\usepackage{comment}
\usepackage{breqn}
\usepackage[nolist,nohyperlinks]{acronym}
\usepackage{wrapfig,lipsum}
\usepackage{graphicx}
\usepackage{epstopdf}
\usepackage{rotating}
\usepackage{amsmath, amssymb, amsthm}
\usepackage{stmaryrd}
\usepackage{tikz}
\usepackage{hyperref}
\usepackage{verbatim}
\usepackage{url}
\usepackage[utf8]{inputenc}
\usepackage[english]{babel}

\usepackage{multicol}
\usepackage{xr-hyper}

\usepackage{scalerel}
\usepackage{multicol}

\usepackage[noadjust]{cite}
\usepackage[normalem]{ulem}
\usepackage{color}
\usepackage{dsfont}
\usepackage{bm}
\usepackage[short]{optidef} 
\usepackage{diagbox}
\usepackage{enumitem}
\usepackage{slashbox}
\usepackage[ruled]{algorithm2e}
\usepackage{multirow}
\usepackage[T1]{}
\usepackage{color, colortbl}
\usepackage{babel}
\usepackage{verbatim}
\usepackage{textcomp}
\usepackage{tabulary}
\usepackage{booktabs}
\usepackage{caption}
\usepackage[numbers]{natbib}
\definecolor{Gray}{gray}{0.95}
\definecolor{LightCyan}{rgb}{0.8,0.85,1}
\definecolor{LightBlue}{rgb}{0.6,0.6,1}
\usepackage[font=small]{caption}
\usepackage{mathtools}
\usepackage{pgfplots}
\pgfplotsset{compat=1.18}

\newcommand{\xhdr}[1]{\vspace{1mm} \noindent {\bf #1.}}

\setlist{nosep}

\usepackage{mathtools}

\newcommand{\frameworkname}[0]{Hermes}
\definecolor{niceblue}{HTML}{007ED6}

\begin{document}

\title{Hermes: A Large Language Model Framework on the Journey to Autonomous Networks}
\author{
Fadhel Ayed$^{\dagger}$, Ali Maatouk$^{\dagger +}$, Nicola Piovesan$^\dagger$, Antonio De Domenico$^\dagger$, Merouane Debbah$^\ddagger$, Zhi-Quan Luo$^\mathsection$\\
$^\dagger$Paris Research Center, Huawei Technologies, Boulogne-Billancourt, France
\\ $^\ddagger$Khalifa University of Science and Technology, Abu Dhabi, UAE
\\ $^\mathsection$The Chinese University of Hong Kong, Shenzhen, China
\\ $^+$Yale University, New Haven, Connecticut, USA\vspace{-15pt}}
\maketitle
\thispagestyle{empty}

\begin{abstract}
The drive toward automating cellular network operations has grown with the increasing complexity of these systems. Despite advancements, full autonomy currently remains out of reach due to reliance on human intervention for modeling network behaviors and defining policies to meet target requirements. Network Digital Twins (NDTs) have shown promise in enhancing network intelligence, but the successful implementation of this technology is constrained by use case-specific architectures, limiting its role in advancing network autonomy. A more capable network intelligence, or ``telecommunications brain", is needed to enable seamless, autonomous management of cellular network.
Large Language Models (LLMs) have emerged as potential enablers for this vision but face challenges in network modeling, especially in reasoning and handling diverse data types. To address these gaps, we introduce Hermes, a chain of LLM agents that uses ``blueprints" for constructing NDT instances through structured and explainable logical steps. Hermes allows automatic, reliable, and accurate network modeling of diverse use cases and configurations, thus marking progress toward fully autonomous network operations.
\end{abstract}

\section{Introduction}
\label{sec:intro}
Since the inception of large-scale cellular networks, researchers and the industry have aimed to automate their operation and management due to the costly and extensive human labor involved. However, communication systems are notorious for their dynamic nature, spanning from wireless channel and network load to unpredictable faults and errors that require network adaptation. Due to these characteristics, full network autonomy has not yet been achieved, and human presence in the operation loop is still prevalent at multiple levels of the hierarchical network stack. On the network autonomy scale~\cite{autonomouslevels}, where at level 0 humans perform network operations entirely using manual procedures and level 5 refers to full network automation, current network operations lie in the middle, aiming to reach level 3 automation before 2026~\cite{currentnetworks}. 
\ac{NDT} has emerged as a highly promising candidate to enhance the design, analysis, operation, automation, and intelligence of future mobile networks~\cite{SRCON2023Intro}. 
However, the impact of this technology toward full network autonomy is limited by the current design approach of \acp{NDT} where different use cases are mapped to distinct \ac{NDT} architectures~\cite{oranDT}, as processing different types of data and modeling and optimizing distinct network functionalities within a unique piece of software is extremely complex.

To break this barrier and take autonomy beyond this level, a more capable type of network intelligence is needed, which encompasses extensive knowledge about network operations and functionalities, and can streamline these operations
– essentially functioning as a \textit{telecommunications brain}.
The notion of a telecommunications brain has been utilized in the literature before to refer to a large-scale intelligent entity capable of understanding the intricacies of the network, the various cause-and-effect relationships in its functionalities, and the ability to plan and predict the network behavior in advance. This intelligence continuously monitors the network state, promptly reacts to any unforeseen changes, and seamlessly adapts the network operations to new scenarios as they arise. Arriving at such a realm of telecommunications brain is an ambitious goal that would elevate network autonomy to new heights. Although the end goal is clear, the path to realizing a telecommunications brain is still an open challenge.

Then, \acp{LLM} have emerged, revolutionizing the \ac{AI} field, especially \ac{NLP}, by propelling text generation, comprehension, and interaction to unprecedented levels of sophistication. Promptly, researchers in the network domain have identified \acp{LLM} as key enablers to pave the way to the telecommunications brain~\cite{bariah2023large}. Particularly, researchers envisioned a realm where \acp{LLM} take over the driving seat of network operations and management. However, to this day, the application of \acp{LLM} in the telecommunications domain has been mostly successful as human add-ons, such as \ac{RAG} systems~\cite{bornea2024telco} that fetches \ac{3GPP} standards information and conversational chatbot tools for wireless communication specifications~\cite{kotaru2023adapting}. There have also been recent successful implementations of \acp{LLM} in embedding network commands, such as setting the transmit power of a base station to a specific value, into actionable configuration files for network management~\cite{10097683}. In \cite{mongaillard2024large}, the authors propose an LLM-based multi-agent framework designed to convert user requests into optimized power scheduling vectors by selecting from a set of predefined equations and solvers the most suitable for the intended task. These lines of work leverage the alignment between such translation tasks and the natural language proficiency of \acp{LLM}. In another line of work, to overcome the limitations of \acp{LLM} in the network domain, researchers advocated for multi-modal \acp{LLM} trained on wireless signals and network measurements to augment their capabilities \cite{xu2024large}. However, creating such large multi-modal models presents significant challenges. These challenges include the need for extensive datasets of measurements and wireless signals, often proprietary to operators and vendors, the complexity of integrating diverse modalities, and the inherent weakness of \acp{LLM} in managing numerical operations and relationships ~\cite{maatouk2023large}. Therefore, despite the current research hype around \acp{LLM}, the question remains open: do \acp{LLM} truly hold the key for achieving the so-called telecommunications brain and leading to full autonomy in telecommunication networks?

This work aims to address this question through a multi-step approach. As a first step, we posit that, fundamentally, an \ac{LLM} can be considered inching closer to becoming a telecommunications brain if it can grasp the causal relationships between network components, configurations, parameters, and their impact on network performance. In other words, this capability would allow \acp{LLM} to construct end-to-end network models-- \acp{NDT}-- to effectively handle new environments and unseen circumstances. 

Moreover, we demonstrate that, today, the most powerful \acp{LLM}, e.g., GPT-4o
, struggle to perform well on understanding and modeling the network behavior, even when using advanced prompting techniques like chain-of-thought~\cite{wei2023chainofthoughtpromptingelicitsreasoning}. We shed light on the common pitfalls \acp{LLM} fall into and the typical mistakes they make. Our findings illustrate that despite their impressive capabilities, current \acp{LLM} are far from being autonomous agents capable of taking the driving seat for telecommunications network management and operations on their own. 

To address these pitfalls, we introduce Hermes: a comprehensive chain-of-agents \ac{LLM} framework that tackles network modeling and automation through the elaboration of ``blueprints" of \acp{NDT}. In this context, a blueprint is a set of step-by-step logical blocks autonomously designed and coded by the LLMs using their parametric knowledge of the telecommunications domain, rather than relying on the direct interpretation of network measurements as multi-modal models do \cite{xu2024large}. By incorporating key components such as self-reflection steps and feedback mechanisms, along with a granular step-by-step logical approach, Hermes ensures the validity of these blueprints and their associated code to realize a \ac{NDT} tailored to the tasked intent.

We demonstrate how leveraging the blueprints of \ac{NDT} significantly increases the reliability of the \ac{LLM} in addressing diverse network modeling tasks, resulting in a more robust comprehension of network dynamics and operations.

\section{LLMs as Key Enabler for Autonomous Networks}
\label{sec:llmsnetop}

\begin{figure}[t]
    \centering
    \includegraphics[width = \linewidth]{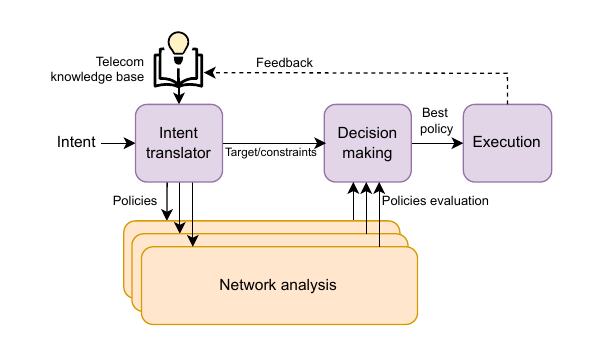}
    \caption{{Policy deployment in autonomous networks.}}
    \label{fig:overview}
\end{figure}

Autonomous Network tasks involve the adjustment of network parameters, network planning, and anomaly resolution. Effectively solving these tasks requires a fundamental understanding of how network parameters interact, how functionalities interconnect, and how various configurations affect overall performance. To achieve the highest level of network autonomy \cite{autonomouslevels}, this well-grounded understanding is essential.

Figure~\ref{fig:overview} illustrates the key stages of a policy deployment in autonomous networks. The process begins with the formulation of an \textit{intent}, a high-level objective, such as ``reduce network energy consumption by 2\%".
The intent is processed by the \textit{intent translator}, which leverages a telecommunications knowledge base enriched with historical data and domain expertise. The translator converts the abstract intent into a set of candidate policies—actionable strategies that could fulfill the defined objective. Additionally, it defines target \acp{KPI} and constraints. 
For example, a candidate policy might involve activating a shutdown mechanism in certain base stations. Next, the candidate policies are evaluated using a network analysis framework, such as a \ac{NDT}, which assesses each policy based on multiple \acp{KPI}, including energy efficiency, latency, and throughput. The evaluation results are then passed to the decision-making module, where the policies are ranked according to targets and constraints defined by the intent translator. 
The most effective policy is selected and sent to the execution phase for implementation in the network.
After execution, the system collects performance feedback, which is used to refine the telecommunications knowledge base, enabling continuous learning and enhancement. 

The success of the executed policy in achieving the original intent is determined by the accuracy of the network analysis stage, which, in turn, depends on the availability of precise network models. These models must also be flexible enough to simulate the effects of any candidate policy.

Today, policy formulation and evaluation rely heavily on human experts who use diverse types of network data to characterize network behavior and develop rule-based strategies to meet high-level intents. This expert-driven approach, while effective for a limited set of policies, is constrained by the cost of measurement campaigns and manual intervention, making it unsustainable for large-scale future networks. Moreover, system-level simulations, another common approach for network analysis, often fail to accurately model the capabilities of real network products and tend to be generic, lacking specificity to the actual network in question.

Given these limitations, there is growing interest in \acp{NDT}, which combine expert knowledge with \ac{ML} and network data to realize advanced solutions for modeling the network~\cite{SRCON2023Intro}. However, the current design of \acp{NDT} suffers from limited scalability and reusability. Different use cases and \acp{MNO} typically focus on distinct KPIs. Moreover, data availability may vary across service providers and regions, e.g., due to distinct regulations. As a result, researchers and engineers have to frequently design new \ac{NDT} architectures to meet evolving requirements \cite{oranDT}. In the following sections, we show how \acp{LLM} can address these challenges by integrating their internal knowledge with available network data and expert-designed models.

\subsection{Modeling Requirements for Autonomous Networks}
To identify the capability of a \ac{NDT} to manage network operations, we analyze autonomous network tasks that need a clear understanding of causal relationships in network behavior. Integrating these capabilities into an \ac{LLM} framework contrasts with typical tasks that focus on translating intent to configuration files or chatbot applications, which primarily leverage, together with \ac{NLP} capabilities, the \ac{LLM} superficial knowledge of telecommunications. Instead, autonomous network tasks demand a deeper comprehension of network causal relationships and intelligent decision-making in applying this understanding.

\xhdr{1. Network Parameters Optimization} This task requires \acp{LLM} to identify which network parameters to control and the algorithm to apply for their dynamic adjustment, all while achieving desired outcomes without negatively affecting the network performance. For instance, selecting the transmission modulation and coding scheme that maximizes the spectral efficiency without exceeding the acceptable packet error rate illustrates an understanding of how optimization parameters influence \acp{KPI} within the network. 

\xhdr{2. Network Policy Recommendation} This task requires the \ac{LLM} to grasp the intricate cause-and-effect relationships among various network functionalities and components, going beyond merely understanding how network parameters influence \acp{KPI}. The \ac{LLM} must model the implications of implementing specific policies on network functions. For example, activating an energy efficiency solution based on base station shutdown can impact user mobility and resource allocation within the network.

\xhdr{3. Network Planning} This task requires the \ac{LLM} to navigate unknown network scenarios by leveraging its predictive capabilities and the ability to generalize its internal knowledge. 
One example of this task is deciding on the optimal location for installing a new site to enhance network capacity and/or coverage while preserving the \ac{SINR} in nearby cells.

Given the complexity of these tasks, it is challenging to envision a single \ac{LLM} excelling in all of them, as we will further illustrate in later sections. To address this challenge, we propose a modular framework composed of a chain of LLM agents, specifically designed to excel across these diverse autonomous network tasks.

\subsection{Why LLMs struggle in modeling network behavior?}
\label{sec:LLMstruggle}
\acp{LLM} offer sparks of intelligence, but their capabilities need to be effectively supported by human knowledge to successfully achieve the targeted task. In the following, we discuss the key limitations of \acp{LLM} in modeling tasks, which have guided our choices of framework design.

\textbf{Performing computations:} Recent research shows that \acp{LLM} struggle with data manipulation \cite{maatouk2023large}; even the strongest models, such as GPT-4o, often make errors in simple tasks like assessing which of two numbers, such as 9.11 and 9.9, is the largest. 
To address these limitations, we argue that \acp{LLM} should focus on reasoning and code generation, while a dedicated code interpreter handles all computations and data manipulations.

\textbf{Knowledge:} State-of-the-art \acp{LLM} exhibit limitations in their knowledge of the telecommunication domain. This issue is particularly pronounced in smaller models, such as Phi-2 and LLaMA 3, which demonstrate a restricted understanding, especially in relation to telecommunication standards-specific details~\cite{maatouk2023large}. A notable approach to enhance \ac{LLM} performance in telecommunications is \ac{RAG} \cite{bornea2024telco}, which integrates external knowledge bases to significantly improve domain-specific expertise.

\textbf{Planning:} 
\acp{LLM} are known for their limited ability to perform effective planning, which is an area of ongoing research~\cite{valmeekam2023planning}. Complex tasks, such as those requiring the consideration of multiple levels of abstraction --as in autonomous networks-- demand that the \ac{LLM} maintains a broad, high level perspective while simultaneously managing precise details for each individual logical step. Despite advancements in the field, state-of-the-art \acp{LLM} frequently exhibit deficiencies in this context. For instance, \acp{LLM} may overemphasise on individual logical steps at the expense of the overall task context, as well as being unable to account for all relevant constraints and dependencies.

\textbf{Concept to execution:} A common issue with \acp{LLM} is their difficulty in translating conceptual knowledge into correct execution. Even when they grasp a concept, they often struggle to apply the elementary reasoning steps that humans instinctively use to adapt general ideas to specific situations. An example of this challenge arises in tasks involving mathematical concepts. For instance, while an \ac{LLM} may understand the concept of \ac{SINR} and know how to compute it, it may fail to execute the task correctly due to errors in handling measurement units. For example, it might apply the linear \ac{SINR} formula while using input values in dBm.

\textbf{Hallucinations:} \acp{LLM} may generate plausible-sounding but inaccurate or entirely fabricated information, commonly referred to as hallucinations. This occurs because \acp{LLM} are designed to predict the next word based on patterns in their training data, without fact-checking or verification. Even when using reasoning strategies like chain-of-thought, they can still produce incorrect results. Moreover, \acp{LLM} struggle to recognize these hallucinations, highlighting the need for external feedback mechanisms.

\section{Hermes Framework}
\label{sec:methodology}
\begin{figure*}[h]
    \centering
    \includegraphics[width = \linewidth]{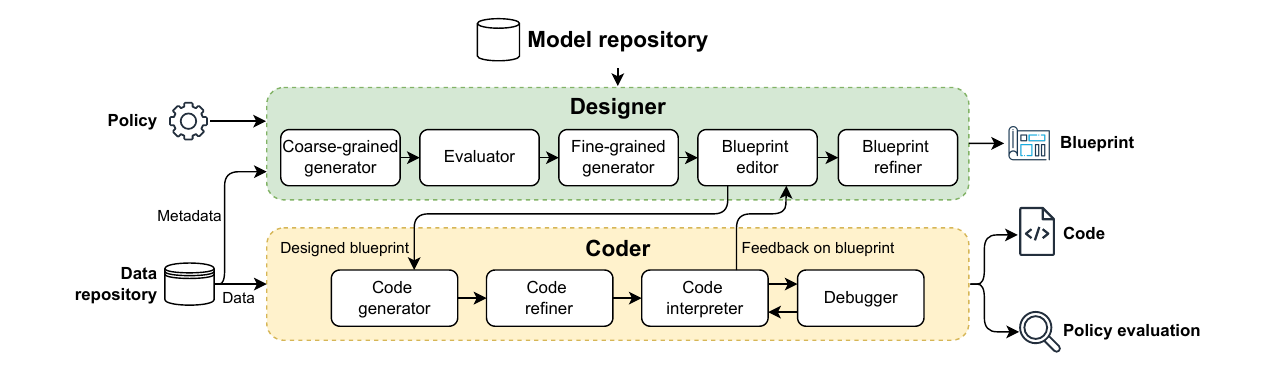}
    \caption{Architecture of the \frameworkname{} Framework.}
    \label{Architecture}
\end{figure*}

In this section, we introduce \frameworkname{}, a chain-of-agents framework that uses ``blueprints'' for constructing \acp{NDT} instances able to automatically analyze the impact of any policy implemented in the network.
\frameworkname{} separates the network modeling tasks into two roles: Designer and Coder. 
The Designer interprets the candidate policy proposed by the intent translator and develops a modeling strategy to assess its impact on network \acp{KPI}, utilizing the available network data for evaluation.  We denote this strategy a \textit{blueprint}; the blueprint consists of the necessary sequence of models with corresponding formulas and the related explanations to build a \ac{NDT} (see Figure \ref{fig:enter-label}). 
The Coder takes the blueprint as input and implements it as a Python program. Using \ac{LLM} capabilities and a Python interpreter, the Coder writes and executes the code, addressing potential bugs.

The Designer and Coder work together in a three-phase process. First, the Designer drafts an initial blueprint. Next, this blueprint is translated into executable code by the Coder. In the final phase, a feedback loop is established: the Designer iteratively refines the blueprint based on the results of the numerical evaluation performed by the Coder on test data, while the Coder integrates these updates and ensures the code remains error-free. 
This framework ultimately delivers a numerical evaluation of the candidate policy, along with a detailed blueprint outlining the steps taken and the corresponding Python code that implements the blueprint.

To address the \ac{LLM} limitations discussed in Sec. \ref{sec:LLMstruggle}, specific techniques are employed. To enhance planning capabilities, \frameworkname{} utilizes a multi-scale approach inspired by LLM-based coding agents \cite{ridnik2024code}, beginning with a coarse-grained strategy to capture high-level aspects and followed by iterative refinements. Hallucinations are mitigated through multiple feedback mechanisms: generation agents are complemented by validation agents that employ the Foresee and Reflect framework \cite{zhou2023far}, prompting the \acp{LLM} to anticipate potential issues and reflect on proposed solutions, complementing the quantitative feedback obtained from the code execution.

\subsection{Design Phase}
The overarching objective of this phase is to formulate a comprehensive solution for the network modeling task before proceeding to the coding phase. To accomplish this, the \ac{LLM} agents interact across multiple levels of abstraction, as detailed below:

\textbf{Initial Reflections:}
A group of $N$ LLM agents, denoted as \textit{coarse-grained generators}, takes as input the network modeling task and a description of the available data. Afterward, each agent is prompted to generate several high-level reflections regarding the given task. These reflections consist of conceptual understandings of the task, devoid of detailed equations or implementation details. They may include, for example, identifying the quantities to be calculated or the incremental steps necessary to address the task at hand.

\textbf{Validation of Reflections:}
The original task, along with the reflections generated by each coarse-grained generator, is fed as input to the corresponding next agent in the chain, the \textit{evaluator}. The evaluators validate these reflections and mitigate potential confirmation bias, probing deeper into the ideas presented. They identify challenges and propose solutions, addressing issues such as data unavailability and inaccuracies related to the task. This feedback is essential for ensuring the robustness of the subsequent steps.

\textbf{Fine-Grained Generation:}
After validation, a group of $M$ \textit{fine-grained generators} synthesize the successful outputs from the evaluators and coarse-grained generators. This process mirrors genetic algorithms, where these generators combine insights from various ``parent" reflections to produce a more refined ``child" output. These fine-grained generators generate a comprehensive strategy with associated mathematical formulas and/or pseudo-code.

\textbf{Blueprint Creation and Refinement:}
The refined strategies are passed to the \textit{blueprint editor}, which synthesizes them and constructs the initial blueprint. A blueprint is defined as a YAML file detailing the steps required to accomplish the task, specifying each step’s name, required inputs, produced outputs, and underlying logic.

After the initial blueprint is created, it undergoes further refinement by the \textit{blueprint refiner}. This agent identifies and addresses potential issues, such as missing terms, validates formulated equations, and confirms the appropriateness and scale of the quantities involved in the calculations. 

Upon completing this refinement process, the blueprint can be implemented. The focus then shifts to the next phase, where the blueprint is translated into executable code.
\subsection{Coding Phase}
At the end of the design phase, we obtain a YAML file detailing the necessary logic required to devise an executable code for the blueprint. However, directly transforming the YAML file into executable code by an \ac{LLM} is inherently unreliable. Therefore, mirroring the approach taken in the design phase, we employ an iterative procedure to translate the blueprint logic into executable code, as detailed below:

\textbf{Initial Code:} 
Based on the blueprint, the first agent in the coding chain, the \textit{code generator}, generates a Python implementation reflecting the logic presented in the blueprint. This initial code then undergoes verification by another LLM agent, \textit{the code refiner}, employing a collaborative approach similar to the previous design phase. When prompted to verify the code, the code refiner is also provided with a set of frequently encountered issues that Hermes has observed, such as ensuring the use of appropriate scales for the involved quantities, maintaining the correct data frame column order, and ensuring the necessary files are loaded before executing the blueprint logic.

\textbf{Tracebacks Debugging:} 
After the refinement stage, the resulting code is executed by a Python interpreter at the \textit{code interpreter}. If the execution completes successfully, the process continues with the feedback phase.
However, in the event of tracebacks (error messages), the code interpreter forwards these errors to the \textit{debugger}, which iteratively refines the code until it executes successfully. If, after multiple iterations, the code still fails to run, the entire process is restarted from the beginning.

\subsection{Feedback Phase}
With an executable code in place, the final phase begins, focusing on performing sanity checks on the generated blueprint. At this stage, the blueprint is structured as a YAML file comprising several blocks, each associated with an executable code. To proceed with the sanity checks, we distinguish between two types of blocks:
\begin{itemize}
\item \textbf{Operation blocks:} These are sets of logic that implement specific network strategies or adjustments, such as turning off a cell or modifying parameters.
\item \textbf{Functional blocks:} These contains logic that produces quantitative outputs based on given inputs, such as calculating the \ac{SINR} or determining cell association from \ac{RSRP} levels.
\end{itemize}
Sanity checks can be performed on functional blocks independently of the complete blueprint pipeline. To this end, in \frameworkname{}, the code interpreter executes the code associated with these blocks using sample inputs from the data repository and then compares the outputs against available ground truth. The results are fed back to the designer, which reviews and refines the blueprint if necessary.
This step allows to detect and correct mismatches in unit handling and flaws in equations, ensuring the accuracy of the blueprint before further execution. 

\section{Experimental Results}
\label{sec:experimentalresults}
In this section, we evaluate the performance of Hermes across a series of autonomous network tasks and configurations. We generate the ground truth data for each scenario using a numerical simulator modeling a small network composed by 10 tri-sectored \acp{BS}. 

\begin{figure}
    \centering
    \includegraphics[scale=0.8]{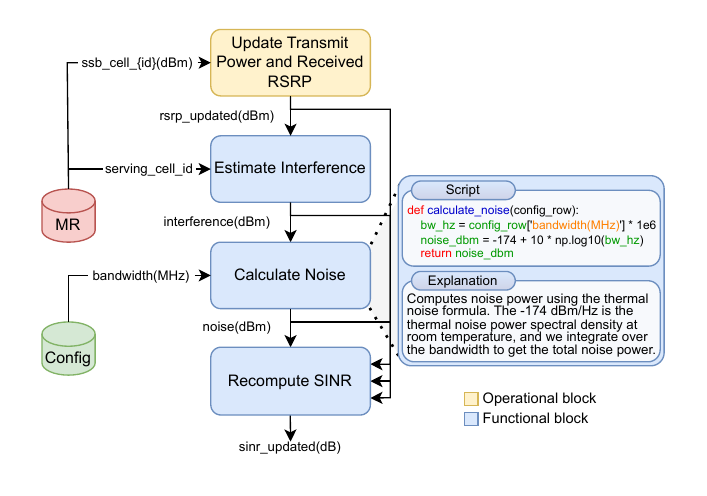}
    \caption{Example of a blueprint designed by \frameworkname{} for the power control task.}
    \label{fig:enter-label}
\end{figure}
\subsection{Hermes performance for different tasks}
In this section, we assess the performance of Hermes in solving four distinct modeling problems, each characterized by varying degrees of complexity. The specific tasks are described as follows:

\begin{itemize}
\item \textbf{Power control:} This use case involves a power control mechanism where the transmit power is dynamically adjusted over time. Hermes must estimate the resulting impact of these power variations on the \ac{SINR} experienced by the \acp{UE}. This task requires at least 4 modeling blocks.
\item \textbf{Energy saving:} Here, one of the BS in the network is turned off to save energy. Hermes is tasked with modeling the effect of this event on the overall network energy consumption. This task requires at least 5 modeling blocks.
\item \textbf{Energy saving vs SINR:} Here Hermes must evaluate the effect of the BS shutdown on the \ac{SINR} at the \ac{UE} level, thereby addressing both energy savings and \ac{QoS} simultaneously. This task requires at least 6 modeling blocks.
\item \textbf{New BS deployment:} This task involves the deployment of a new \ac{BS} in the network. Hermes must estimate the new \ac{SINR} of all the \acp{UE} under the new network configuration. This task requires at least 7 modeling blocks.
\end{itemize}

Each of these problems presents a different level of difficulty. For instance, to solve the first use case, and understand how the variation in transmit power affects the \ac{SINR} at the \acp{UE}, Hermes must model how the power adjustments influence the \ac{RSRP} at the \acp{UE} from the serving \acp{BS}, estimate the interference levels based on the \acp{RSRP} of surrounding \acp{BS}, and finally compute the SINR by incorporating the thermal noise at the receiver. This problem involves a relatively straightforward analysis of power and interference relationships. In contrast, the more complex use cases require additional modeling components, to account, for instance, for deactivated BSs and energy consumption. As an example, we described one instance of the input/output process of Hermes for the power control task in Figure \ref{fig:enter-label}.

We employed GPT-4o as the LLM for all Hermes agents and evaluated the entire framework across 20 independent runs for each of the four tasks.
The results of the generated blueprints are classified as correct if the related estimation errors with respect to the ground truth are below 10\%.
The proportion of successful outcomes across these trials is referred to as the \textit{success rate}.
Moreover, we benchmarked the success rate achieved by Hermes against two alternative approaches: 1) CoT, in which GPT-4o utilizes chain-of-thought reasoning to solve the problems \cite{wei2023chainofthoughtpromptingelicitsreasoning} and generates the corresponding code implementation; and 2) Hermes-coder, where the chain-of-thought solution provided by GPT-4o is implemented by the coder block of Hermes, omitting the use of the blueprint designer block.

\begin{figure}
    \centering
    \begin{tikzpicture}
\begin{axis}[
    width=8cm,
    xlabel={Number of Blocks},
    ylabel={Success Rate (\%)},
    legend style={at={(0.5,1.12)},anchor=north,legend columns=3,draw=none,font=\small},
    ymajorgrids=true,
    xmajorgrids=true,
    grid style=dashed,
    xtick={4,5,6,7}, 
    xticklabels={4\\{\small Power \\control}, 5\\{\small Energy \\saving}, 6\\{\small Energy saving vs SINR}, 7\\{\small New BS deployment}},
    x tick label style={align=center, text width=2cm}, 
    ymin=0, ymax=100, 
]

\addplot[dashed,mark=triangle, mark options={solid,fill=green}, mark size=4pt, green, thick,color={rgb,255:red,130;green,179;blue,102}] 
    coordinates {
        (4, 35) 
        (5, 15) 
        (6, 5)
        (7, 5)
    };
\addplot[dashed,mark=square, mark options={solid, fill=blue}, mark size=3pt, blue, thick,color={rgb,255:red,108;green,142;blue,191}] 
    coordinates {
        (4, 65) 
        (5, 45) 
        (6, 30)
        (7, 25)
    };
\addplot[mark=o, mark size=3pt, mark options={fill=red}, red, thick, color={rgb,255:red,184;green,84;blue,80}] 
    coordinates {
        (4, 85) 
        (5, 80) 
        (6, 80)
        (7, 75)
    };

\node[anchor=south] at (axis cs:4,35) {\small 35\%};
\node[anchor=north] at (axis cs:5,25) {\small 15\%};
\node[anchor=north] at (axis cs:6,14) {\small 5\%};
\node[anchor=north] at (axis cs:7,14) {\small 5\%};

\node[anchor=south] at (axis cs:4,65) {\small 65\%};
\node[anchor=north] at (axis cs:5,55) {\small 45\%};
\node[anchor=south] at (axis cs:6,31) {\small 30\%};
\node[anchor=north] at (axis cs:7,35) {\small 25\%};

\node[anchor=south] at (axis cs:4,85) {\small 85\%};
\node[anchor=south] at (axis cs:5,80) {\small 80\%};
\node[anchor=south] at (axis cs:6,80) {\small 80\%};
\node[anchor=north] at (axis cs:7,85) {\small 75\%};

\legend{CoT, Hermes-coder, Hermes}
\end{axis}
\end{tikzpicture}
    \caption{Success rate achieved by different solutions at different problem complexity.}
    \label{fig:complexity}
\end{figure}
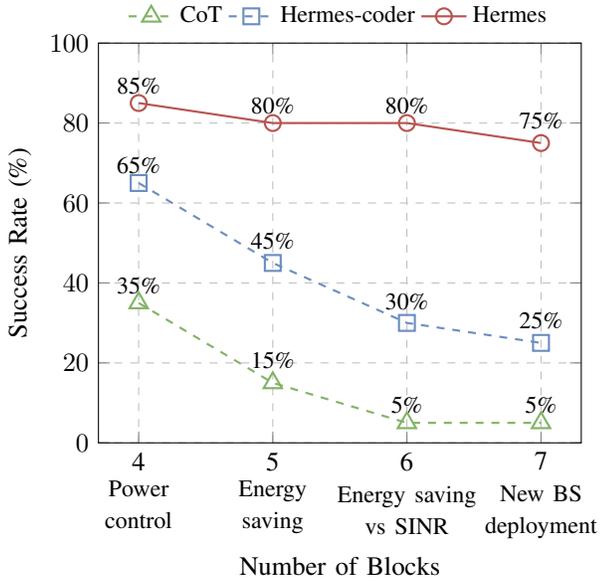

Figure~\ref{fig:complexity} presents the success rate of each method for the four tasks. 
In the simplest task, CoT produced successful estimations 35\% of the time, while Hermes-coder achieved a higher success rate of 65\%. Hermes outperformed both, with an 85\% success rate.
As the complexity of the tasks increased, the performance gap widened. For the most complex task (new cell deployment), CoT’s success rate dropped to 5\%, while Hermes-coder demonstrated significantly better results with a 25\% success rate. However, Hermes consistently delivered the highest performance, maintaining a 75\% success rate.

\subsection{Hermes capabilities with distinct LLMs}
\begin{table}[]
\centering
\caption{Success score of different LLMs on power control and energy saving task.}
\begin{tabular}{lccc}
\toprule
 & \textbf{CoT} & \textbf{Hermes-coder} & \textbf{Hermes} \\
\midrule
\textbf{Llama-3.1-70b} & 0\% & 5\% & 25\% \\
\textbf{Llama-3.1-405b} & 5\% & 15\% & 45\% \\
\textbf{GPT-4o} & 25\% & 55\% & 82.5\% \\
\bottomrule
\label{tab:models_comparison}
\end{tabular}
\end{table}

In this section, we focus on two of the tasks (Power control and Energy saving), and we compare the performance of Hermes and the two benchmarks, Hermes-coder and CoT, when adopting different LLMs. Specifically, we consider two versions of Llama-3.1 (70 billion and 405 billion parameters) and GPT-4o, comparing each configuration by averaging the results over 20 independent runs.

As shown in Table~\ref{tab:models_comparison}, the success-score performance of Hermes and the two benchmarks varies significantly depending on the size and architecture of the LLM used. When using Llama-3.1-70b, the CoT approach produces no successful outcomes, while Hermes-coder shows marginal improvement, achieving a 5\% success rate. However, when using the full Hermes pipeline, performance improves to 25\%. The larger \mbox{Llama-3.1-405b} model yields better results across all approaches, with CoT achieving a 5\% success rate, Hermes-coder increasing to 15\%, and the full pipeline reaching 45\%.

In contrast, GPT-4o significantly outperforms both versions of Llama-3.1 in all configurations. The CoT approach, when combined with GPT-4o, results in a 25\% success rate, while Hermes-coder achieves 55\%, demonstrating the positive impact of its code generation and refinement capabilities. The full Hermes pipeline, leveraging GPT-4o, delivers the best performance, with a success rate of 82.5\%.

The limited performance achieved by the tested open-source models can be attributed to three primary factors.
First, the prompts in Hermes are optimized for GPT-4o's capabilities. It is well known that the performance of LLMs strongly depends on prompt formulation, with different models responding better to specific prompts. Adapting these GPT-4o-tailored prompts could notably improve performance when using alternative, open-source models; Second, although open-source models are advancing and narrowing the gap with proprietary models, significant disparities persist, especially in complex planning and reasoning tasks. 
Third, open-source models show lower proficiency in wireless modeling, frequently generating oversimplified models and demonstrating inferior coding abilities compared to their closed counterparts.

\subsection{Open-source LLMs empowered by expert-designed blocks}\label{expe-blocks}

As mentioned, the results from the previous section indicate that open-source models with limited size lack the knowledge required to accurately design white-box models for wireless networks, resulting in poor performance. However, although these LLMs may not be capable of independently constructing white-box models, they may still be proficient enough to design blueprints using a pre-existing repository of expert-designed white-box models (see the top of Figure \ref{Architecture}).

To explore this, we evaluated two versions of Llama-3.1 --70 billion and 405 billion parameters-- when provided with access to a repository of expert-designed models. This repository contains a varying numbers of pre-designed models that can be selected by the designer and integrated into the blueprint. As in the previous section, we focus on two task: power control and energy saving, and we average the results obtained in 20 independent runs.

Table~\ref{tab:expertblocks} presents the success rates achieved by \mbox{Llama-3.1-70b} and \mbox{Llama-3.1-405b} as the number of expert-designed models in the repository increases.
With Llama-3.1-70b, Hermes achieves a 25\% success rate when no expert-designed models are available. However, performance improves steadily as additional models are included, reaching a 75\% success rate with five expert-designed models.

Similarly, Llama-3.1-405b shows notable performance gains, starting at 45\% without any expert models and rising to 80\% with the integration of five expert-designed models.

\begin{table}[]
\centering
\caption{Performance with varying numbers of expert-designed blocks}
\label{tab:expertblocks}
\begin{tabular}{lcccccc}
\toprule
\textbf{} & \multicolumn{6}{c}{\textbf{Number of expert-designed blocks}} \\
\cmidrule(lr){2-7}
\textbf{} & \textbf{0} & \textbf{1} & \textbf{2} & \textbf{3} & \textbf{4} & \textbf{5} \\
\midrule
\textbf{Llama-3.1-70b} & 25\% & 25\% & 30\% & 45\% & 60\% & 75\%\\
\textbf{Llama-3.1-405b} & 45\% & 50\% & 65\% & 70\% & 75\% & 80\%\\
\bottomrule
\end{tabular}
\end{table}

\section{Future axes of research}
Building on the insights gained from our experiments, several key areas have emerged that could further enhance the capabilities and accuracy of our framework.
Our findings highlight the critical role of human-designed models in enhancing the performance of our framework, emphasizing the need for a well-structured codebase of fundamental components, incorporating data-driven NDTs~\cite{SRCON2023Intro}, system-level simulators, and other white-box models. Establishing a systematic storage and efficient retrieval mechanism for these elements could greatly improve the accuracy and operational efficiency of our framework.

Moreover, maintaining a repository of successful blueprints would enable Hermes to learn from previous experiences, facilitating the development of more complex solutions by reusing components or even entire blueprints. This approach aligns with the principles of curriculum learning, wherein accumulated knowledge is leveraged for solving progressively harder tasks.

The integration of real-time measurement data offers another promising avenue for improvement. However, efficiently managing the large data volumes --potentially reaching terabytes per hour at a city-wide scale-- poses a considerable challenge. Therefore, developing strategies for the efficient storage and processing of representative data will be critical to optimizing the framework capabilities.

\section{Conclusion}
\label{sec:conclusion}
In this paper, we introduced Hermes, a chain of LLMs that employs structured ``blueprints" to construct NDT instances through clear and explainable logical steps. This innovative approach facilitates automatic, reliable, and accurate network modeling across a variety of use cases and configurations, representing a significant advancement toward fully autonomous network operations.
Our findings demonstrate that utilizing NDT blueprints markedly enhances the reliability of the LLMs when tackling diverse network modeling tasks, leading to a more comprehensive understanding of network dynamics and operations. Importantly, Hermes achieved a notable accuracy of up to 80\% in these tasks.

\bibliographystyle{IEEEtran}

\bibliography{reference.bib}

\begin{acronym}[AAAAAAAAA]
  \acro{3GPP}{Third Generation Partnership Project}
 \acro{AI} {Artificial Intelligence}
 \acro{BERT}{Bidirectional Encoder Representations from Transformers}
  \acro{DL}{Deep Learning}
  \acro{DT}{Digital Twin}
   \acro{FPGA}{Field-programmable gate array}
  \acro{GPT}{Generative Pre-trained Transformer}
  \acro{KPI}{Key Performance Indicator}
 \acro{LLM}{Large Language Model}
\acro{BS}{Base Station}
  \acro{ML}{Machine Learning}
  \acro{MNO}{Mobile Network Operator}
 \acro{NLP}{Natural Language Processing} 
   \acro{NDT}{Network Digital Twin}
\acro{NN}{Neural Network}
 \acro{RAN}{Radio Access Network}
\acro{LoRA}{Low-rank adaptation}
\acro{RAG}{Retrieval Augmented Generation}
\acro{RSRP}{Reference Signal Received Power}
\acro{SNR}{Signal-to-Noise Ratio}
\acro{SINR}{Signal-to-Noise plus Interference Ratio}
\acro{QoS}{Quality of Service}
\acro{API}{Application Programming Interface}
\acro{UE}{User Equipment}
 \end{acronym}

\end{document}